\documentclass[conference]{IEEEtran}
\IEEEoverridecommandlockouts

\usepackage{cite}
\usepackage{amsmath,amssymb,amsfonts}
\usepackage{graphicx}
\usepackage{textcomp}
\usepackage{xcolor}
\usepackage{multirow}
\usepackage{multicol}
\usepackage{caption}
\RequirePackage{algorithm}
\usepackage{algpseudocode}
\usepackage{booktabs}
\usepackage{multirow}

\def\BibTeX{{\rm B\kern-.05em{\sc i\kern-.025em b}\kern-.08em
    T\kern-.1667em\lower.7ex\hbox{E}\kern-.125emX}}

\begin{document}


\title{Diversity-Aware Client Selection for Communication Efficient Federated Learning}





\author{
\IEEEauthorblockN{
Max Halleryd\IEEEauthorrefmark{1},\IEEEauthorrefmark{2} 
Sima Sinaei\IEEEauthorrefmark{1} 
Viktor Valadi\IEEEauthorrefmark{3} 
Mina Alibeigi\IEEEauthorrefmark{4}
Henrik Abrahamsson\IEEEauthorrefmark{1}
Salman Toor\IEEEauthorrefmark{3},\IEEEauthorrefmark{5}
}

\IEEEauthorblockA{
\IEEEauthorrefmark{1} \textit{RISE Research Institutes of Sweden}, Stockholm \
\IEEEauthorrefmark{2} \textit{KTH Royal Institute of Technology}, Stockholm \\
\IEEEauthorrefmark{3} \textit{Scaleout Systems}, Uppsala \
\IEEEauthorrefmark{4} \textit{Zenseact}, Gothenburg \
\IEEEauthorrefmark{5} \textit{Uppsala University}, Uppsala \
}
}

\maketitle

\begin{abstract}

Federated learning (FL) enables collaborative model training across distributed clients without sharing raw data, making it attractive for privacy-sensitive and bandwidth-constrained applications. A practical challenge in deploying FL is reducing communication overhead while maintaining model performance under heterogeneous, non-IID data distributions. Selecting which subset of clients participates in each training round offers a principled way to address both challenges simultaneously, yet its impact on data diversity and representativeness remains underexplored. In this paper, we investigate three client selection strategies, Power-of-Choice, Fisher information, and Centered Kernel Alignment, evaluating them not only by accuracy and convergence, but through the lens of participation diversity using the Gini coefficient and KL divergence. Non-IID conditions demonstrate that restricting participation to $15\%$ of clients per round achieves significant communication savings, whereas when selection promotes diverse and balanced contributions, it also improves generalization over biased alternatives. This work provides a systematic and reproducible analysis demonstrating that diversity-focused client selection renders communication efficiency and model quality complementary rather than competing objectives.

\end{abstract}


\section{Introduction}

Federated learning (FL) \cite{mcmahan2017communication} is a distributed machine learning paradigm in which a global model is trained collaboratively across multiple clients, each holding its own local dataset. Unlike centralized learning, FL keeps data decentralized and aggregates only model updates, making it well suited for privacy-sensitive and bandwidth-constrained applications ranging from hospitals and banks to IoT networks and smart manufacturing \cite{hasan2025federated}.


Despite its advantages, FL introduces challenges that do not arise in centralized settings. Among these is communication overhead: since model updates must be transmitted repeatedly across training rounds, communication becomes a bottleneck in large-scale systems or under limited network bandwidth. This is particularly acute in vehicular and edge networks, where connectivity is intermittent and latency requirements are strict \cite{jia2025comprehensive}. A second fundamental challenge is data heterogeneity \cite{barona2025heterogeneity}. In practice, client datasets are rarely independent and identically distributed (non-IID), as differences in user behavior, geography, or data collection processes introduce significant variation in local distributions. This heterogeneity can bias local updates, slow convergence, and degrade the generalization of the global model.


Client selection, choosing which subset of clients participates in each training round, offers a principled way to address both challenges simultaneously, yet existing work evaluates it primarily in terms of convergence speed or final accuracy. The question of \textit{which} clients are selected is equally consequential: under non-IID conditions, biased selection can systematically exclude underrepresented clients or classes, quietly degrading generalization even as communication costs fall. Diversity-aware selection strategies, by contrast, have the potential to maintain or improve model quality under strict participation constraints, though this perspective has received limited attention.


In this paper, we have carefully selected three client selection strategies with the focus on diversity-aware selection, Power-of-Choice\cite{powerofchoice}, Fisher information \cite{martens2020new}, and Centered Kernel Alignment (CKA) \cite{kornblith2019similarity}, and evaluate them not only by accuracy and convergence, but through the lens of participation diversity using the Gini coefficient and KL divergence. Experiments show that diversity-focused selection makes communication efficiency and model quality complementary rather than competing objectives. We provide a systematic and reproducible analysis that offers actionable guidance for deploying FL in bandwidth-constrained, real-world environments.

    \label{fig:fl}

\section{Methodology}
\label{sec:methods}
This section defines the experimental framework used in this study. First, the federated learning setup and communication constraints are introduced, followed by the datasets and their non-IID partitioning. The evaluated client selection strategies are then described, along with the diversity metrics used for analysis. Finally, the model architectures used for each dataset are specified.

\subsection{Federated Learning Setup}
We consider a cross-device federated learning setting with $K=20$ clients. In each communication round, a subset of $m=3$ clients is selected to perform local training, while the baseline uses full participation ($m=20$). The restrictive size of the subset is deliberately chosen as a stress test. All methods use standard federated averaging (FedAvg) for aggregation.

\subsection{Datasets and Data Heterogeneity}
We evaluate on the MNIST handwritten digit dataset and ImageNet50. The latter consists of 50 classes from the larger ImageNet-1K image classification dataset. For both datasets, an 80/20 split is used to create local training and test sets. 

To simulate realistic non-IID conditions, data is partitioned across clients using a Dirichlet distribution \cite{hsu2019measuring} with parameter $\alpha = 0.5$, inducing moderate label skew. This results in heterogeneous local datasets with varying class distributions and dataset sizes.

\subsection{Client Selection Strategies}
At each round, a subset of clients is selected based on one of the following criteria:


\textbf{Power-of-choice:} Selects clients with the highest local loss, prioritizing difficult and underrepresented data while implicitly encouraging diverse participation across training rounds. In the experimental setup, $d = 10$ clients are sampled with probability proportional to local dataset size $|\mathcal{D}_i|$:
\begin{equation}
    p_i = \frac{|\mathcal{D}_i|}{\sum_{j=1}^{K} |\mathcal{D}_j|}
\end{equation}
and the $m = 3$ clients with the highest local loss are selected.


\textbf{Fisher information:} Clients are ranked by the trace of the empirical Fisher information matrix, approximated from local gradients:
\begin{equation}
    \mathrm{Tr}(\hat{F}_i) = \frac{1}{|\mathcal{D}_i|} \sum_{x \in \mathcal{D}_i} \left\| \nabla_\theta \ell(\theta; x) \right\|^2_2
\end{equation}
Where $\nabla_\theta \ell(\theta; x)$ is the gradient of the loss function, $\theta$ is the parameters of the model and $\| \cdot\|_2$ denotes the $\ell_2$ norm. Clients with higher Fisher trace are selected, corresponding to updates with higher parameter sensitivity.


\textbf{CKA-based selection:} Centered Kernel Alignment (CKA) measures the similarity between representations of the local and global models. For representation matrices $X \in \mathbb{R}^{n \times p}$ and $Y \in \mathbb{R}^{n \times q}$, linear CKA is defined as:
\begin{equation}
    \text{CKA}(X, Y) = \frac{\left\| Y^\top X \right\|_F^2}{\left\| X^\top X \right\|_F \left\| Y^\top Y \right\|_F}
\end{equation}
where $\|\cdot\|_F$ denotes the Frobenius norm and $n$ is the number of samples used to compute the representations. Clients are selected based on their CKA score, prioritizing low similarity.

Here we would like to highlight, one obvious candidate for client selection is \textbf{Random selection}. This strategy was intentionally excluded from this study. Since it produces non-deterministic results, it cannot be reliably compared across experiments or guaranteed to be reproducible.

\subsection{Diversity Metrics}
To analyze the impact of client selection, we quantify diversity along two dimensions:


\textbf{Selection inequality:} Measured using the Gini coefficient \cite{cowell2010gini} over client participation counts:
\begin{equation}
    G = \frac{\sum_{i=1}^{K} \sum_{j=1}^{K} |f_i - f_j|}{2K \sum_{i=1}^{K} f_i}
\end{equation}
where $f_i$ is the number of times client $i$ is selected. Lower Gini values indicate more uniform client selection.


\textbf{Dataset coverage:} Measured using the KL divergence \cite{cover2006elements} between the label distribution of selected clients $P$ and a uniform distribution $U$:
\begin{equation}
    D_{\text{KL}}(P \| U) = \sum_{c=1}^{C} P(c) \log \frac{P(c)}{U(c)}
\end{equation}
where $C$ is the number of classes and $U(c) = \frac{1}{C}$. Lower divergence indicates more balanced class representation.

\subsection{Models}
For MNIST, a lightweight CNN with two convolutional layers, two pooling layers and two fully connected layers is used. For ImageNet50, we use ConvNeXt-T due to its efficiency and strong performance on image classification tasks \cite{convnext}.

\section{Experimental Results and Discussion}
\label{Ex}
All experiments use partial participation with 3 out of 20 clients per round, yielding a communication reduction of $6.46\times$ (including two warmup rounds with full participation). We focus on how \emph{client selection diversity} impacts convergence and generalization.

\subsection{The MNIST Handwritten Digit Dataset}


MNIST provides a controlled setting in which the effect of client selection on diversity metrics can be isolated and precisely evaluated, free from the confounding effects of dataset complexity (Table~\ref{tab:mnist_acc}). All methods achieve performance close to baseline accuracy except CKA, whose similarity-based selection criterion leads to biased client participation and consequently negatively impact on accuracy. The differentiating factor shifts from whether the model learns to the efficiency and uniformity of the learning process itself. Here, the total upstream data sent by the baseline is $288.79$ MB, while the partial participation approaches only send $44.72$ MB, i.e. a reduction of $244.07$ MB by using partial participation.


\begin{table}[h]
\centering
\caption{MNIST accuracy (\%) averaged over last 15 rounds.}
\begin{tabular}{l|c|c}
\hline
Strategy & Train Acc. & Test Acc. \\ \hline
Baseline & 95.48 & 94.95 \\
CKA      & 79.67 & 76.41 \\
Fisher   & 93.19 & 91.04 \\
Power-of-choice & 93.51 & 91.76 \\
\hline
\end{tabular}
\label{tab:mnist_acc}
\end{table}

\subsubsection{Selection Diversity}
\begin{figure}[h]
    \centering
    \includegraphics[width=0.7\linewidth]{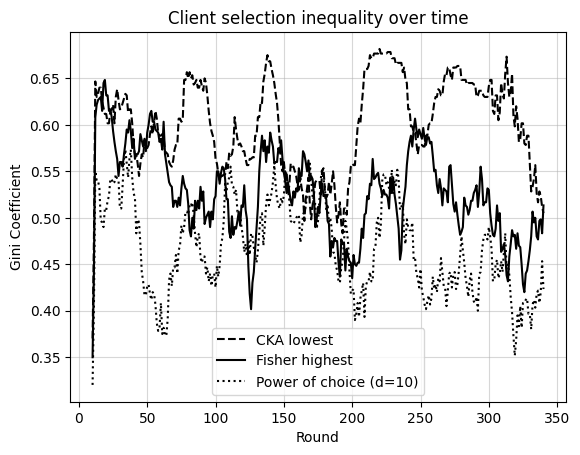}
    \caption{Gini coefficient (MNIST) over $350$ rounds, where $3$ clients are selected in each round.}
    \label{fig:mnist_gini}
\end{figure}

Lower Gini coefficients (Fig.~\ref{fig:mnist_gini}) correlate with improved generalization. Power-of-choice achieves the lowest mean inequality $(0.47)$ and highest test accuracy among partial methods, while CKA exhibits the highest mean inequality $(0.59)$ and worst performance. This indicates that diverse client participation improves generalization, consistent with findings in prior work.     
\begin{figure}[h]
    \centering
    \includegraphics[width=0.7\linewidth]{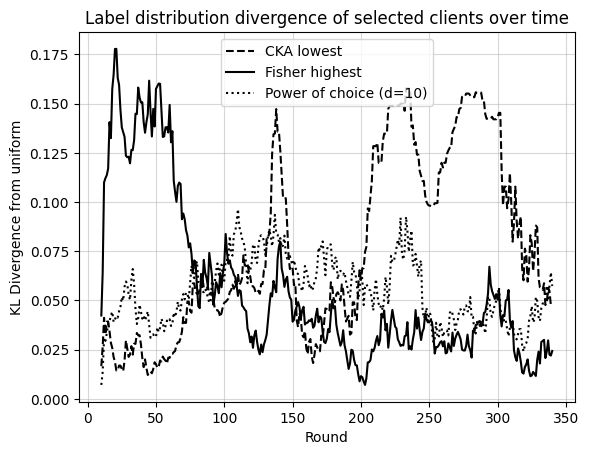}
    \caption{KL divergence to the uniform distribution (MNIST); $3$ clients selected per round.}
    \label{fig:mnist_kl}
\end{figure}
KL divergence (Fig.~\ref{fig:mnist_kl}) further supports this: Fisher and power-of-choice converge to low divergence, indicating balanced class coverage. In contrast, CKA maintains higher divergence, reflecting biased sampling and resulting in unstable training and reduced accuracy.

Together, Figs. \ref{fig:mnist_gini}  and \ref{fig:mnist_kl} present a coherent picture: Power-of-Choice (with restrictive participation to only $15\%$ of clients) achieves the lowest participation inequality, the most balanced class coverage, and the highest accuracy among partial methods, not independently, but as a direct consequence of its loss-guided selection naturally rotating across underrepresented clients.
\begin{figure}[H]
    \centering
    \vspace{-10pt}
    \includegraphics[width=0.7\linewidth]{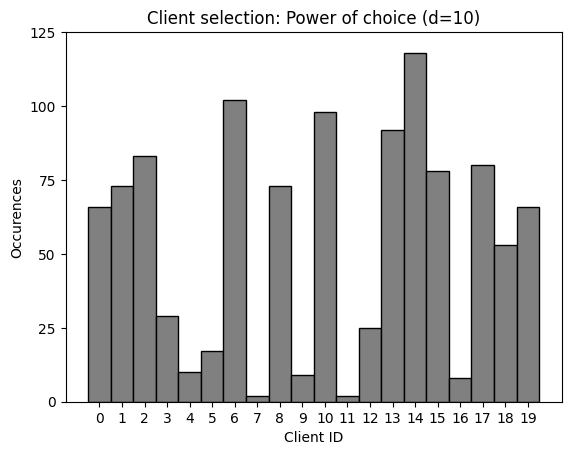}
    \caption{Histogram of clients selected during training for Power-of-Choice (MNIST)}
    \label{fig:mnist hist}
\end{figure}
Power-of-choice does not result in a small number of clients dominating the selection (Fig \ref{fig:mnist hist}). However, it exhibits near-complete exclusion of certain clients, which are primarily selected only during the initial full-participation warmup rounds. These clients tend to have more limited class diversity in their local datasets; for example, client $7$ contains five classes and client $11$ contains four. This reduced class coverage is associated with lower local training loss, which in turn decreases their likelihood of being selected under the power-of-choice criterion. As a result, some clients contribute minimally during later stages of training, which influences the overall representation of classes in the aggregated model.


\subsection{ImageNet50 Dataset}

ImageNet50 is deliberately chosen as a stress test. With greater class complexity and non-IID skew, the weaknesses of each selection strategy become amplified, making this dataset the more informative of the two. In this experiment, the total upstream data sent by the baseline is $780.05$ GB, while the other approaches send $120.79$ GB, i.e. a reduction of $659.25$ GB by using partial participation.


\begin{table}[htp]
\centering
\caption{ImageNet50 accuracy (\%) averaged over last 15 rounds.}
\begin{tabular}{l|c|c}
\hline
Strategy & Train Acc. & Test Acc. \\ \hline
Baseline & 97.89 & 59.77 \\
CKA      & 87.90 & 51.84 \\
Fisher   & 93.01 & 52.92 \\
Power-of-choice & 91.77 & 53.22 \\
\hline
\end{tabular}
\label{tab:imgnet_acc}
\vspace{-10pt}
\end{table}

\subsubsection{Selection Diversity}
\begin{figure}[htp]
    \vspace{-5pt}
    \centering
    \includegraphics[width=0.7\linewidth]{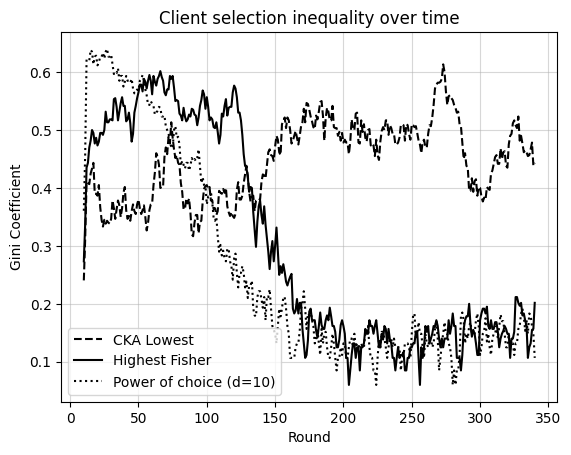}
    \caption{Gini coefficient (ImageNet50) over $350$ rounds, where $3$ clients are selected in each round.}
    \label{fig:imgnet_gini}
\end{figure}

Figure~\ref{fig:imgnet_gini} shows that all methods start with high inequality, selecting a small subset of clients in early training. Fisher and power-of-choice progressively reduce inequality, approaching uniform participation, while CKA remains biased. This transition aligns with improved convergence stability.

\begin{figure}[htp]
    \centering
    \includegraphics[width=0.7\linewidth]{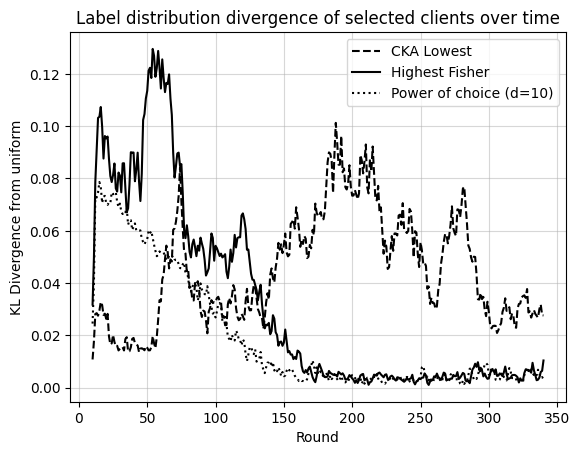}
    \caption{KL divergence to the uniform distribution (ImageNet50); $3$ clients selected per round.}
    \label{fig:imgnet_kl}
\end{figure}

KL divergence trends (Fig.~\ref{fig:imgnet_kl}) reveal that early training is dominated by limited class coverage (high divergence). As training progresses, Fisher and power-of-choice reduce divergence significantly, indicating improved dataset representativeness. CKA maintains higher divergence, suggesting persistent class imbalance, which explains its lower generalization performance.

The accuracy gap between baseline and Power-of-Choice ($59.77$ vs. $53.22$) should be read in context: only $15\%$ of clients participate per round, data is both non-IID and imbalanced, $50$ labels, and no additional regularization is applied. Under less conservative participation rates of $50\%$, this gap is narrow while communication savings remain significant. Our recent experiments under less conservative participation rates confirm this, with results available on request. The presented results should therefore be interpreted as a characterization of worst-case constraints rather than a fundamental limitation of the approach.

\begin{figure}[h]
    \vspace{-5pt}
    \centering
    \includegraphics[width=0.7\linewidth]{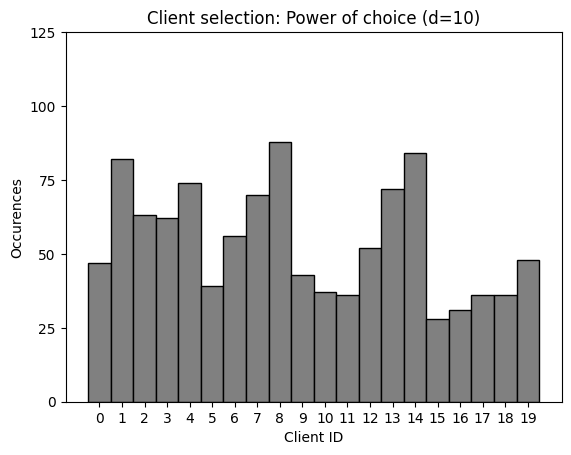}
    \caption{Histogram of clients selected during training for Power-of-Choice (ImageNet50)}
    \label{fig:imagenet hist}
\end{figure}
Clients $8$, $13$, and $14$ have the smallest local dataset sizes in the ImageNet50 setting, yet they are among the most frequently selected under the Power-of-Choice strategy (Fig \ref{fig:imagenet hist}). These clients also exhibit relatively high local training loss. This indicates that, despite sampling candidates with probability proportional to dataset size, the final selection step based on local loss can favor clients with smaller datasets, leading to more fair selection. 


\subsection{Key Insights}

Across both datasets, the results reveal a consistent pattern: strategies that promote diverse client participation achieve better generalization under strict communication constraints. The Gini coefficient and KL divergence serve not merely as descriptive metrics but as predictive indicators. Power-of-Choice maintains the lowest participation inequality and most balanced class coverage, directly corresponding to superior accuracy among partial-participation methods.

Critically, these results are achieved with only $15\%$ client participation per round, yielding a  $6.46\times$ communication reduction while preserving competitive accuracy. The joint analysis of selection diversity and communication efficiency at this participation rate has not been previously reported, and the findings provide actionable guidance for deploying FL in bandwidth-constrained environments.


\section{Conclusion and Future Work}

This paper investigated the impact of client selection on communication efficiency, convergence, and generalization in federated learning under non-IID conditions. By evaluating Power-of-Choice, Fisher information, and CKA through the lens of participation diversity, we showed that diversity-focused selection renders efficiency and model quality complementary rather than competing objectives, achieving a $6.46\times$ communication reduction at only $15\%$ client participation while preserving competitive accuracy. The results further highlight that dataset complexity significantly moderates the effect of selection strategies, and that conclusions drawn from simple benchmarks should be interpreted with caution. Future work should explore adaptive selection mechanisms that explicitly account for data coverage, and evaluate these strategies on more complex, real-world datasets under stronger non-IID conditions.


\vspace{-2pt}
\section{Acknowledgement}
This work was carried out within the DREAM project (2025-01024), funded by Sweden's Innovation Agency (Vinnova) under the Advanced Digitalization initiative.

\bibliographystyle{IEEEtran}
\bibliography{references.bib}

\end{document}